\documentclass{article}

\usepackage[numbers]{natbib}
\usepackage[preprint,nonatbib]{neurips_2020}

\usepackage[utf8]{inputenc}
\usepackage[T1]{fontenc}
\usepackage{nicefrac}
\usepackage{microtype}
\usepackage{placeins}

\usepackage{fullpage}
\usepackage{amsmath,amsthm,amsfonts}
\usepackage{amssymb}
\usepackage{mathrsfs}
\usepackage{mathtools}
\usepackage{stmaryrd} 
\usepackage{colonequals}
\usepackage{xifthen}

\usepackage{graphicx}
\usepackage{multirow}
\usepackage{booktabs}
\usepackage[inline]{enumitem}
\usepackage{caption}
\usepackage{subcaption}
\usepackage[dvipsnames]{xcolor}
\usepackage{tikz,pgfplots}

\usepackage[colorlinks = true]{hyperref}

\hypersetup{
	urlcolor  = NavyBlue,  
	linkcolor = cyan,      
	citecolor = purple,    
	filecolor = magenta    
}

\usepackage{cancel}

\usepackage{thmtools}
\newtheoremstyle
	{custom}      
	{\topsep}     
	{\topsep}     
	{\normalfont} 
	{}            
	{\bfseries}   
	{}            
	{\newline}    
	{%
		\rule{\textwidth}{0.4pt} \\*%
		\thmname{{#1}}~\thmnumber{{#2}}\thmnote{~(\textbf{#3})}. \\[-1.5ex]%
		\rule{\textwidth}{0.4pt}%
	}             

\theoremstyle{custom}

\newcommand{\mb}[1]{\boldsymbol{\mathbf{#1}}}
\newcommand{\mc}[1]{\mathcal{#1}}

\newcommand{\mbb}[1]{\mathbb{#1}}

\DeclarePairedDelimiter\roundbracket{(}{)}
\DeclarePairedDelimiter\squarebracket{[}{]}
\DeclarePairedDelimiter\curlybracket{\{}{\}}
\makeatletter
\def\rbr{\@ifnextchar[{\roundbracket}{\roundbracket*}}
\def\sbr{\@ifnextchar[{\squarebracket}{\squarebracket*}}
\def\cbr{\@ifnextchar[{\curlybracket}{\curlybracket*}}
\makeatother

\newcommand{\seq}[1]{\left[ {#1} \right]}

\newcommand{\expo}[1]{\exp \left( {#1} \right)}

\newcommand{\R}{\mbb{R}}

\usepackage{amsmath,amsfonts,bm}

\def\rvh{{\mathbf{h}}}

\def\rvx{{\mathbf{x}}}

\DeclareMathAlphabet{\mathsfit}{\encodingdefault}{\sfdefault}{m}{sl}
\SetMathAlphabet{\mathsfit}{bold}{\encodingdefault}{\sfdefault}{bx}{n}

\newcommand{\mask}{\texttt{[mask]}~}
\newcommand{\cls}{\texttt{[cls]}~}
\newcommand{\name}{Funnel-Transformer~}

\definecolor{mygreen}{HTML}{15b01a}
\definecolor{myred}{HTML}{be0119}

\newcommand{\green}[1]{{\color{mygreen}#1}}

\title{Funnel-Transformer: Filtering out Sequential Redundancy for Efficient Language Processing}

\author{
	Zihang Dai$^{*12}$, Guokun Lai$^{*1}$,  Yiming Yang$^1$,  Quoc V. Le$^2$\\
	$^1$Carnegie Mellon University, $^2$Google AI Brain Team \\
	\texttt{\{dzihang,guokun,yiming\}@cs.cmu.edu, qvl@google.com}
}

\begin{document}

\maketitle

\renewcommand{\thefootnote}{\fnsymbol{footnote}}
\footnotetext[1]{Equal contribution.}
\renewcommand{\thefootnote}{\arabic{footnote}}

\begin{abstract}
With the success of language pretraining, it is highly desirable to develop more efficient architectures of good scalability that can exploit the abundant unlabeled data at a lower cost.
To improve the efficiency, we examine the much-overlooked redundancy in maintaining a full-length token-level presentation, especially for tasks that only require a single-vector presentation of the sequence.
With this intuition, we propose Funnel-Transformer which gradually compresses the sequence of hidden states to a shorter one and hence reduces the computation cost.
More importantly, by re-investing the saved FLOPs from length reduction in constructing a deeper or wider model, we further improve the model capacity.
In addition, to perform token-level predictions as required by common pretraining objectives, Funnel-Transformer is able to recover a deep representation for each token from the reduced hidden sequence via a decoder.
Empirically, with comparable or fewer FLOPs, Funnel-Transformer outperforms the standard Transformer on a wide variety of sequence-level prediction tasks, including text classification, language understanding, and reading comprehension.\footnote{The code and pretrained checkpoints are available at \footnotesize \url{ github.com/laiguokun/Funnel-Transformer}.}
\end{abstract}

\section{Introduction}
\label{sec:intro}
With the recent success of unsupervised language pretraining~\cite{peters2018deep,devlin2018bert,yang2019xlnet,liu2019roberta,clark2020electra,lan2019albert,kong2019mutual,raffel2019exploring,lewis2019bart,song2019mass,liu2019multi,song2020mpnet}, the power of neural self-attention models (a.k.a. Transformer)~\cite{vaswani2017attention} has been pushed to a new level, leading to dramatic advancements in machine learning and 
natural language processing (NLP).
More importantly, it has been observed that with more FLOPs invested in longer pretraining and/or larger models, the performance of pretrained Transformer models consistently improve.
However, it is extremely expensive to pretrain or even just to finetune the state-of-the-art self-attention models, as they require much more FLOPs and memory resources compared to traditional models in NLP.
This largely limits their applications and success in more fields.

Given this challenge, there has been an increasing amount of efforts to reduce the costs of pretraining and finetuning self-attention models.
From the perspective of post-pretraining processing, typical approaches include distillation, pruning and quantization of various kinds, which try to derive a lighter model from an well-pretrained model by taking advantage of the richer signals in the larger model or learning to remove less important operations.
Another line of research aims at designing an architecture that not only has a lower resource-to-performance ratio (more efficient) but also \textit{scales as well as} the Transformer, at least in certain domains.
Most of such methods build upon the Transformer backbone and focus on redesigning its building blocks. 
Representative solutions include searching for better micro operation or macro module designs~\cite{so2019evolved,chen2020adabert}, replacing the full pairwise attention with local operations such as convolution~\cite{wu2020lite} and dynamic convolution~\cite{wu2019pay}, and optimizing the hidden size combinations for existing blocks~\cite{sun2020mobilebert}.

Across the wide variety of ideas mentioned above, a common strategy is to identify redundant operations or representations and replace them with more efficient ones.
Inspired by this line of thinking, in this work, we will be focusing on the potential redundancy induced by always maintaining a \textit{full-length sequence} of hidden representations across all layers in Transformer.
Intuitively, for many sequence-level NLP tasks such as text classification and ranking, the most common use case is to extract a \textit{single} vector from the entire sequence, which does not necessarily preserve all information down to the token-level granularity.
Hence, for such tasks, the full-length sequence of hidden states may contain significant redundancy.
This is analogous to the case of image recognition, where the convolution neural network gradually reduces the spatial resolution/size of feature maps as the neural network goes deeper.
In addition, linguistic prior also encourages gradually merging nearby tokens (words) into larger semantic units (phrases), which naturally leads to a shorter sequence of representations.

Concretely, we propose to gradually reduce the sequential resolution (i.e. length) of the hidden representation in self-attention models.
Immediately, the reduction in sequence length can lead to significant savings in both FLOPs and memory.
More importantly, the saved computational resource can be directly re-invested in constructing a deeper (or wider) model to boost the model capacity without additional computational burden.
In addition, to address the challenge that common pretraining objectives such as masked language modeling (MLM)~\cite{devlin2018bert} require separate representations for each token, we design a simple strategy to decode a full-length sequence of deep representations from the hidden state of reduced length.
As a result, the proposed model can be directly trained without modifying the pretraining objectives, as well as adopted for downstream tasks that require token-level representations.

Empirically, with comparable or even fewer FLOPs, by trading sequential resolution for depth, our proposed model achieves an improved performance over the standard Transformer on a wide variety of sequence-level prediction tasks, including text classification, language understanding, and reading comprehension.

\section{Method}
\label{sec:method}

\subsection{Background}
\label{sec:background}

\paragraph{Transformer Architecture} The Transformer architecture~\cite{vaswani2017attention} is a highly modularized neural network, where each Transformer layer consists of two sub-modules, namely the multi-head self-attention (S-Attn) and position-wise feed-forward network (P-FFN).
Both sub-modules are wrapped by a residual connection and layer normalization.
Schematically, given a length $T$ sequence of hidden states $\mb{h} = \seq{h_1, \dots, h_T}$, the computation of a single Transformer layer can be expressed as
\begin{align}
\label{eqn:attn}
\mb{h} &\leftarrow \text{LayerNorm}\rbr{\mb{h} + \text{S-Attn}(\text{Q}=\mb{h}, \text{KV}=\mb{h})}, \\
\label{eqn:pffn}
h_i &\leftarrow \text{LayerNorm}\rbr{h_i + \text{P-FFN}(h_i)},\quad \forall i = 1, \cdots, T.
\end{align}
\paragraph{Pretraining Objectives} The most commonly used pretraining objective is the masked language modeling (MLM) proposed by BERT~\cite{devlin2018bert}.
For a length-$T$ natural language sequence $\mb{x}$ sample from a large unlabeled set $\mc{D}$, the MLM objective first constructs a corrupted sequence $\hat{\mb{x}}$ by randomly replacing 15\% of the tokens of $\mb{x}$ with a special token \mask and then trains a Transformer model~\cite{devlin2018bert} to reconstruct the original $\mb{x}$ based on $\hat{\mb{x}}$, i.e.,
\[ 
\max_{\theta}\; \mc{J}_\text{MLM}(\theta) = 
	\mbb{E}_{\mb{x} \sim \mc{D}} \mbb{E}_{\mc{I}} \sum_{i \in \mc{I}} \log P_\theta(x_i \mid \hat{\mb{x}}_\mc{I}) = 
	\mbb{E}_{\mb{x} \sim \mc{D}} \mbb{E}_{\mc{I}} \sum_{i \in \mc{I}} \log 
	\frac{ \expo{ e(x_i)^\top h_{i}\rbr{\hat{\mb{x}}_\mc{I}} } }
		 { \sum_{x'} \expo{ e(x')^\top h_{i}\rbr{\hat{\mb{x}}_\mc{I}} } },
\]
where $\mc{I}$ is the positions of masked tokens, the subscript in $\hat{\mb{x}}_\mc{I}$ emphasizes its dependence on $\mc{I}$, $e(x)$ denotes the embedding of the token $x$, and $h_{i}\rbr{\hat{\mb{x}}_\mc{I}}$ the last-layer hidden state at position $i$ produced by the Transformer model.
After pretraining, the entire model is finetuned in downstream tasks.

To show the generality of our proposed model, we also experiment with another pretraining objective ELECTRA~\cite{clark2020electra}.
Different from MLM, ELECTRA relies a pair of jointly trained generator and discriminator.
Specifically, the generator usually has a smaller size (1/4 of that of the discriminator) and is directly trained via the MLM objective, i.e., $\max_{\theta_G} \mc{J}_\text{MLM}(\theta_G)$.
Then, for each masked position, a token is sampled from the reconstruction distribution of the generator to replace the \mask token and form a new sequence $\tilde{\mb{x}}$, i.e., 
if $i \in \mc{I}$, $\tilde{x}_i \sim P_{\theta_G}(x_i \mid \hat{\mb{x}}_{\mc{I}})$ else $\tilde{x}_i = x_i$.
Given the new sequence $\tilde{\mb{x}}$, the discriminator is then trained to distinguish whether each token in $\tilde{\mb{x}}$ is real (same as $\mb{x}$) or fake (different from $\mb{x}$) via binary classification.
After pretraining, only the discriminator will be used during finetuning and the generator is simply discarded.

\paragraph{Discussion} Note that both pretraining objectives introduced above require the ability to produce a hidden state for each input token, i.e., $h_{i}\rbr{\hat{\mb{x}}_\mc{I}}$ and $h_{i}\rbr{\tilde{\mb{x}}}$.
Due to this requirement, it seems natural to keep a full sequence of hidden states.
However, in contrast, many sequence-level downstream tasks like classification or ranking only need a single-vector summary of the entire sequence.
Fundamentally, this suggests that some kind of compression is usually required to remove the unnecessary redundancy during finetuning.
This observation immediately leads to the following two questions:
\begin{itemize}[leftmargin=*,itemsep=0em,parsep=0em,topsep=0em]
\item Can we design a general model that is equally expressive but more efficient by compressing the full sequence of hidden states into a more compact form?
\item With the compressed representations, how can the model retain the ability to produce token-level representations for pretraining?
\end{itemize}
To answer these two questions, we next present our proposed architecture.

\subsection{Proposed Architecture}
\label{sec:proposed}
\begin{figure}[!h]
\centering
\includegraphics[width=0.925\linewidth]{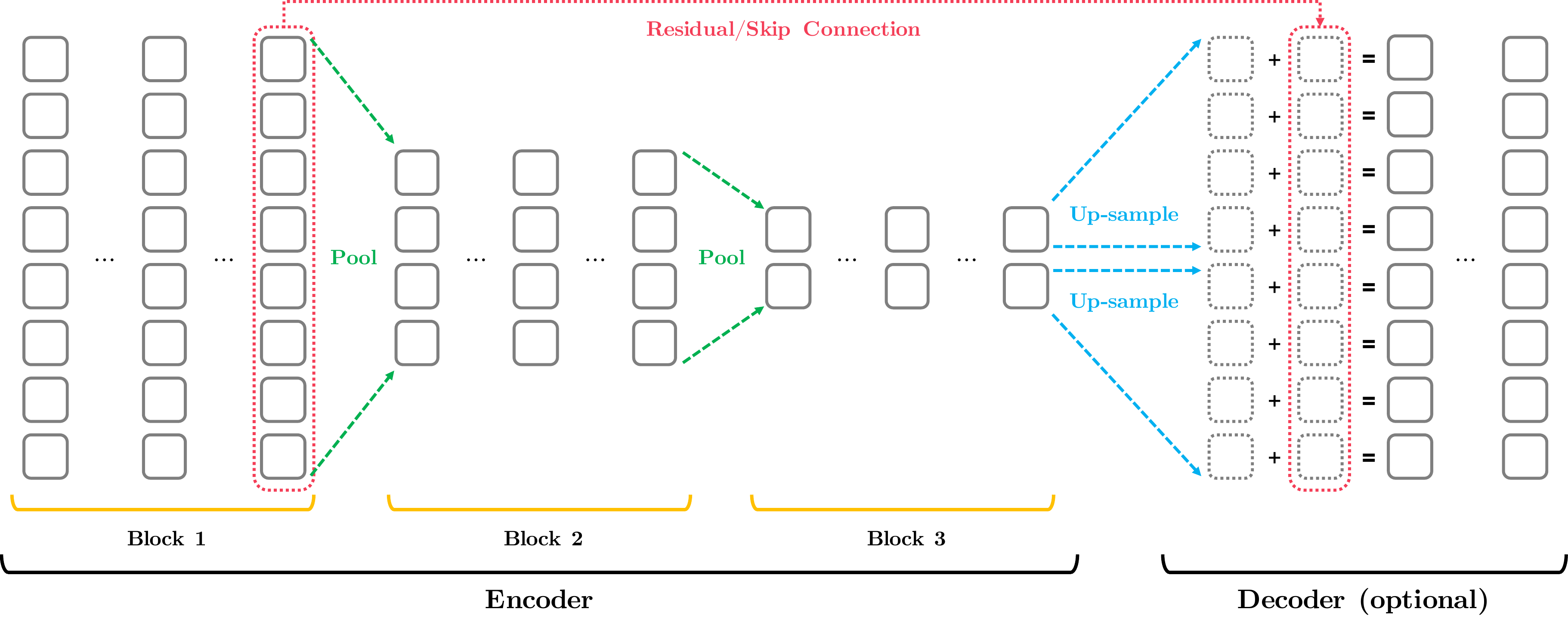}
\caption{High-level visualization of the proposed Funnel-Transformer.}
\label{fig:arch}
\vspace{-0.5em}
\end{figure}
To inherit the high capacity and optimization advantages of the Transformer architecture, the proposed model keeps the same overall skeleton of interleaved S-Attn and P-FFN sub-modules wrapped by residual connection and layer normalization.
But differently, to achieve representation compression and computation reduction, our model employs an encoder that gradually reduces the sequence length of the hidden states as the layer gets deeper.
In addition, for tasks involving per-token predictions like pretraining, a simple decoder is used to reconstruct a full sequence of token-level representations from the compressed encoder output.

\paragraph{Encoder} As illustrated in the left part of Fig.~\ref{fig:arch}, the encoder consists of several blocks of consecutive Transformer layers.
Within each block, the sequence length of the hidden states always remains the same.
But when going from a lower-level block to a higher-level block, the length of the hidden sequence is reduced by performing certain type of pooling along the sequence dimension, i.e.,
\begin{align}
\label{eqn:pooling}
\mb{h}' \leftarrow \text{Pooling}(\mb{h}),
\end{align}
where $\mb{h} \in \R^{T \times D}$ and $\mb{h}' \in \R^{T' \times D}$ for some $T' < T$.
Importantly, instead of directly feeding the pooled sequence $\mb{h}'$ into the first S-Attn layer of the new block, we only use pooled sequence $\mb{h}'$ to construct the query vector (and the residual signal) of the self-attention, while the unpooled sequence $\mb{h}$ serves that role of key and value vectors, i.e.
\begin{align}
\label{eqn:poolq_only_attn}
\mb{h} \leftarrow \text{LayerNorm}\rbr{ \mb{h}' + \text{S-Attn}\rbr{\text{Q}=\mb{h}', \text{KV}=\mb{h}}}.
\end{align}
Note that the output sequence of this special S-Attn module has the same length as the pooled sequence $\mb{h}'$.
To understand the advantage of this particular design, it is helpful to compare the proposed ``pool-query-only'' variant with the naive alternative of using $\mb{h}'$ for both the query and key-value vectors, i.e., $\text{S-Attn}\rbr{\text{Q}=\mb{h}', \text{KV}=\mb{h}'}$:
\begin{itemize}[leftmargin=*,itemsep=0em,topsep=0em]
\item Under the naive approach, the compression is solely controlled by the pooling operation, which is finished before the attention module.
Hence, relatively simple pooling methods such as average/mean pooling won't be able to achieve good compression.
\item Under the pool-query-only variant, the compression depends on not only how the pooling is performed, but also how the self-attention weighted sums the unpooled sequence to form each pooled vector.
Effectively, the particular attention here can be seen as a type of linear compression that combines $T$ bases into a smaller number of $T'$ ``compressed bases''.
Therefore, with minimum computational overhead, this variant makes compression operation more expressive.
\end{itemize}
With this particular pool-query-only design in place, we find the simplest strided mean pooling applied to each sliding window of the sequence work very well in practice.
For simplicity, we only experiment with stride 2 and window size 2 in this work.
Hence, the pooling operation will reduce the sequence by half and each pooled hidden state corresponds to a window of 2 unpooled hidden vectors.
Intuitively, this type of pooling roughly follows the linguistic prior that nearby tokens could be gradually merged (or compressed) into a larger semantic component.
Once the sequence length is halved after the pooling and pool-query-only attention, the rest of the encoder computation simply follows the standard updates in Eqn. \eqref{eqn:pffn} and \eqref{eqn:attn}.

Finally, as an extra implementation detail, recall that a particular design in language pretraining is to add a special token \cls to the beginning of the original input sequence, and use the last-layer hidden state corresponding to \cls (i.e., $h_1$) as the representation of the sequence.
To prevent the pooling from destroying this special structure, we first separate the \cls hidden state and the rest of hidden states and only apply the pooling to the rest of hidden states.
For some practical implementation issues and an efficient solution, we refer readers to Appendix~\ref{sec:appendix-seq-trunc}.

\paragraph{Decoder} In order to recover a full sequence of hidden states from the encoder output of reduced length, a natural idea would be performing some kind of up-sampling.
For instance, in image generation or super-resolution, deconvolution (transposed convolution) or parameter-free resizing with bilinear interpolation are often used to increase the spatial resolution of the feature map.
Hence, we can simply adapt these ideas from 2D processing to our 1D case and apply proper up-sampling to the encoder output.

However, instead of performing multiple up-samplings with small expansion rate (e.g. increasing the sequence length by 2x each time) as in image domain, we here choose to employ a single up-sampling with a large expansion rate, as shown on the right part of Fig.~\ref{fig:arch}.
Specifically, given the output sequence $\mb{h}^\text{M}$ of length $T_M = T / 2^{M-1}$ from an $M$-block encoder, we directly up-sample it to a full-length sequence $\mb{h}^\text{up} = \seq{h^\text{up}_1, \cdots, h^\text{up}_T}$ by repeating each hidden vector $2^{M-1}$ times:\vspace{-0.25em}
\begin{align}
\label{eqn:upsample}
\forall i = 1, \cdots, T,\quad h^\text{up}_i = h^M_{i//2^{M-1}},\vspace{-0.25em}
\end{align}
where $\cdot//\cdot$ denotes floor division.
However, note that every $2^{M-1}$ consecutive vectors in $\mb{h}^\text{up}$ are exactly the same and hence do not contain detailed token-level information.
Hence, we further extract the last-layer hidden states from the \textit{first block} of the encoder $\mb{h}^{1}$, which still has the full length $T$ and contains the uncompressed token-level information.
Then, the lower-level representation $\mb{h}^{1}$ and up-sampled higher-level representation $\mb{h}^\text{up}$ are added together to form a deep token-level representation $\mb{g} = \mb{h}^{1} + \mb{h}^\text{up}$.
Effectively, this forms a residual/skip connection that enables detailed token information and potentially easier optimization.
In addition, we stack a few more Transformer layers upon $\mb{g}$ to achieve a better deep fusion of the low-level and high-level features.
In this work, we always use 2 Transformer layers in decoder.

It is important to emphasize that the decoder is \textit{only used if the task requires token-level prediction}, such as in standard pretraining or sequence labeling.
For tasks that only requires a single vectorial representation of the sequence like classification, the decoder is discarded after pretraining and only the encoder is finetuned.
Finally, to emphasize the filtering/compression property of the encoder as well as its shape, we name the proposed model \texttt{Funnel-Transformer} (F-TFM).

\subsection{Complexity \& Capacity Analysis}
\label{sec:analysis}
With the architecture design specified, we now analyze how the sequence compression affects the complexity and capacity of the proposed model, especially  compared to the standard Transformer.

Firstly, for a Transformer layer with an S-Attn and a P-FFN module of hidden size $D$, the complexity of processing a length-$T$ sequence is $O(T^2D + TD^2)$.\footnote{Since the corresponding memory complexity is simply $O(T^2 + TD)$, which is always offset by a multiplier $1/D$, we will focus on the computation complexity with the conclusion directly carried through.}
Hence, every time the sequence length is reduced by half in the encoder, we enjoy a \textit{super-linear} (more than half) complexity drop.
In practice, as the $O(TD^2)$ term has a large constant, a near-linear speedup is observed more often. 
The super-linear effect is more detectable when the sequence length is relatively long like in pretraining.
Therefore, given the same FLOPs, we can at least trade a full-length layer in the 1st block for $2^{m-1}$ layers in the $m$-th block, which provides an economical way to increase the depth of network.

On the other hand, the capacity of a compressed-length layer is clearly upper-bounded by that of a normal full-length layer.
In most cases where the compression is lossy, reducing the sequence length will inevitably lead to capacity drop.
The good news is that the capacity drop of a single layer could be well compensated by re-investing the saved FLOPs in stacking more cheaper layers of reduced length or increasing the width of the model.

As a concrete example, for a Transformer of BERT$_\text{Base}$ size, i.e., 12 layers of hidden size 768 (L12H768), we may construct a \name of 3 blocks where each block has 6 layers of hidden size 768 (B6-6-6H768).
Despite having 18 layers in total, when finetuned for classification, the FLOPs of the B6-6-6H768 architecture only corresponds to at most $6 + 6 / 2 + 6 / 4 = 10.5$ full-length layers, clearly fewer than that of L12H768.
More importantly, as we will show in the experiments, B6-6-6H768 significantly outperforms L12H768.
While intuitive, how to construct an optimal block layout given this \textit{depth-length trade-off} remains an open challenge.
For this work, we only consider relatively regular layout and leave more systematic studies for future work.

Finally, notice that trading sequential resolution for depth or width has a side effect of increasing the total number of parameters.
For instance, B6-6-6H768 has 1.5x Transformer parameters compared to L12H768.
In practice, more parameters may increase communication cost in distributed training as well as the memory consumption and memory access time.
A simple remedy is to perform certain parameter sharing, as used in ALBERT, to recover the same parameter count.
Taking B6-6-6H768 as an example, one may tie the parameters for every two layers in the 2nd and 3rd blocks, denoted as B6-3x2-3x2H768, which gives back the same number of parameters to L12H768.
However, parameter sharing could result in performance loss.
Fundamentally, this brings us another trade-off between the gain (capacity) and cost (memory and communication cost) of using more parameters, which can be highly device dependent.

\section{Related Work}
\label{sec:related}
As far as we know, no previous work achieves performance gain via compressing the sequence length of the hidden states under language pretraining.
Meanwhile, our proposed model is quite similar to the bottom-up model proposed by a contemporary work~\citep{subramanian2020multi} for causal language modeling.
The key differences include the pool-query-only design for down-sampling, how the up-sampling is performed, and our relative attention parameterization.
Another closely related idea is Power-BERT~\cite{goyal2020power}, which learns to \textit{soft-eliminate} word vectors that are less ``significant'' during finetuning.
Hence, for post-finetuning inference, the sequence length can be reduced to achieve acceleration.
More generally, our work is also related to previous work on hierarchical recurrent neural networks~\cite{lin2015hierarchical} and Transformer models~\cite{zhang2019hibert,garg2019multiresolution}. 
Different from these methods, our model does not rely on any pre-defined hierarchy or boundary of semantic meanings and always captures the full-length dependency input with attention.

In contrast, our work draws many inspirations from the computer vision domain.
The contracting encoder and expanding decoder framework with residual connections is conceptually similar to the ResUNet~\cite{ronneberger2015u} for image segmentation.
The strided pooling is also widely used to construct modern image recognition networks~\cite{scherer2010evaluation}.
Despite the similarities, apart from the obvious difference in data domain and computation modules, our encoder employs a special pool-query-only design to improve the compression, and our decoder only requires a single up-sampling with a large expansion rate.

In addition, a line of research in graph neural networks has tries to gradually reduce the number of nodes in different ways and obtain a single vectorial representation for supervised classification.~\cite{ying2018hierarchical,gao2019graph,lee2019self}
While these methods could potentially be plugged into our model as alternative compression operations, it remains an open question whether compression techniques developed for supervised graph classification can be extended the large-scale language pretraining.
\section{Experiment}
\label{sec:experiment}
In this section, we empirically evaluate the proposed F-TFM by first pretraining it and then finetuning it in downstream tasks.
Following previous work, for pretraining, we consider two common settings:
\begin{itemize}[leftmargin=*,itemsep=0em,parsep=0em,topsep=0em]
\item \textbf{Base scale}: Pretraining models for 1M steps with batch size 256 on Wikipedia + Book Corpus. This is the setting used by original BERT~\cite{devlin2018bert}. 
We will rely on this setting to perform fair comparison between F-TFM and the standard Transformer as well as some ablation studies.
\item \textbf{Large scale}: Pretraining models for 500K steps with batch size 8K on the five datasets used by XLNet~\cite{yang2019xlnet} and ELECTRA~\cite{clark2020electra} (Wikipedia + Book Corpus + ClueWeb + Gigaword + Common Crawl).
We will compare F-TFM trained at this scale with previous state-of-the-art methods.
\end{itemize}
For finetuning, we mainly focus on sequence-level tasks that only requires a single vectorial representation of the input sequence, since F-TFM is designed with such a purpose in mind.
Specifically, such tasks include the GLUE benchmark for language understanding~\cite{wang2018glue}, 7 widely used text (sentiment / topic) classification tasks (IMDB, AD, DBpedia, Yelp-2, Yelp-5, Amazon-2, Amazon-5)~\cite{zhang2015character}, and the RACE reading comprehension dataset~\cite{lai2017race}.
In addition, to see how F-TFM performs when token-level prediction is needed, we consider the SQuAD question answering task which requires the model to select a token span from the context paragraph as the answer.
For more details of the experiment setting, we refer readers to Appendix \ref{sec:appendix-hparam}.

Finally, for all models implemented in this work including Transformer baselines in the base-scale comparison section \ref{sec:exp-base-scale}, we always use the relative positional attention parameterization proposed by Transformer-XL~\cite{dai2019transformer} (see Appendix \ref{sec:appendix-rel-attention} for some implementation details of Transformer-XL).

\subsection{Base-scale Results}
\label{sec:exp-base-scale}
Firstly, we evaluate how F-TFM performs compared to the standard Transformer under similar amount of computation (i.e., FLOPs).
For this purpose, we consider three commonly used model sizes for the standard Transformer, namely large (L24H1024), base (L12H768) and small (L6H768).
Then, for each Transformer baseline, we construct F-TFMs of different block layouts and parameters, while ensuring the F-TFMs always have fewer or similar FLOPs.
Based on the MLM pretraining objective, the results on GLUE benchmark and text classification are presented in Table \ref{tab:fair-comparison-mlm}, where we also include the \textit{relative} FLOPs and \#Params.
\begin{table}[!ht]
\centering
\small
\begin{tabular}{@{}lccccccccc@{}}
	\toprule
	Model size & CoLA & SST-2 & MRPC & STS-B & QQP & MNLI & QNLI & RTE & GLUE-AVG \\
	\cmidrule(r){1-1} \cmidrule(lr){2-9} \cmidrule(l){10-10}
	{L24H1024} 
	& 63.2 & 94.8 & 91.8/88.5 & 91.1 & 88.7/91.7 & 88.7 & 94.0 & 80.5 & 86.6 \\
	{B10-10-10} 
	& 64.8 & 95.0 & 92.5/89.5 & 90.7 & 88.6/91.5 & 88.9 & 94.0 & 81.5 & \bf 87.0 \\
	{B8-8-8} 
	& 63.5 & 94.7 & 92.2/89.0 & 90.7 & 88.9/91.7 & 88.8 & 93.6 & 81.2 & 86.7 \\
	\cmidrule(r){1-1} \cmidrule(lr){2-9} \cmidrule(l){10-10}
	{L12H768} 
	& 60.5 & 93.0 & 92.2/89.0 & 89.4 & 88.1/91.2 & 86.0 & 92.2 & 73.6 & 84.4 \\
	{B6-6-6} 
	& 62.5 & 94.0 & 92.2/89.0 & 89.5 & 88.4/91.4 & 87.0 & 92.7 & 76.5 & \bf 85.3 \\
	{B6-3x2-3x2} 
	& 60.5 & 93.6 & 92.4/89.2 & 89.4 & 88.2/91.3 & 86.4 & 92.5 & 75.0 & 84.7 \\
	{B4-4-4} 
	& 59.1 & 92.7 & 91.8/88.7 & 89.1 & 88.2/91.3 & 85.5 & 92.0 & 73.2 & 83.9 \\
	\cmidrule(r){1-1} \cmidrule(lr){2-9} \cmidrule(l){10-10}
	{L6H768} 
	& 55.2 & 91.5 & 91.1/87.8 & 88.1 & 87.2/90.6 & 82.7 & 90.0 & 64.6 & 81.3 \\
	{B3-4-4} 
	& 59.0 & 92.8 & 91.8/88.5 & 88.5 & 87.8/90.9 & 84.8 & 91.8 & 73.2 & \bf 83.7 \\
	\bottomrule
\end{tabular}

\vspace{0.1em}

\begin{tabular}{@{}lccccccccc@{}}
	\toprule
	Model size & IMDB & AG & DBpedia & Yelp2 & Yelp5 & Amazon2 & Amazon5 & FLOPs & \#Params \\
	\cmidrule(r){1-1} \cmidrule(lr){2-8} \cmidrule(l){9-10}
	{L24H1024} 
	& 4.440 & \bf 4.987 & 0.646 & 1.758 & 28.73 & 2.409 & 32.78 & {1.00x} & {1.00x} \\
	{B10-10-10} 
	& \bf 4.404 & 5.026 & \bf 0.617 & \bf 1.734 & \bf 28.52 & \bf 2.400 & \bf 32.65 & {0.73x} & {1.22x} \\
	{B8-8-8} 
	& 4.552 & 5.079 & 0.664 & 1.713 & 28.84 & 2.438 & 32.87 & {0.58x} & {1.00x} \\
	\cmidrule(r){1-1} \cmidrule(lr){2-8} \cmidrule(l){9-10}
	{L12H768} 
	& 5.328 & 5.184 & 0.663 & 2.013 & 29.35 & 2.571 & 33.14 & {1.00x} & {1.00x} \\
	{B6-6-6} 
	& \bf 4.908 & \bf 5.079 & 0.654 & \bf 1.939 & \bf 29.03 & \bf 2.518 & \bf 32.91 & {0.88x} & {1.39x} \\
	{B6-3x2-3x2} 
	& 5.144 & 5.342 & \bf 0.649 & 1.892 & \bf 29.03 & 2.570 & 33.01 & {0.88x} & {1.00x} \\
	{B4-4-4} 
	& 5.348 & 5.250 & 0.670 & 1.979 & 29.37 & 2.596 & 33.16 & {0.58x} & {1.00x} \\
	\cmidrule(r){1-1} \cmidrule(lr){2-8} \cmidrule(l){9-10}
	{L6H768} 
	& 6.252 & 5.421 & 0.697 & 2.203 & 30.33 & 2.801 & 33.69 & {1.00x} & {1.00x} \\
	{B3-4-4} 
	& \bf 5.520 & \bf 5.342 & \bf 0.670 & \bf 2.042 & \bf 29.51 & \bf 2.603 & \bf 33.16 & {1.00x} & {1.53x} \\
	\bottomrule
\end{tabular}
\caption{
	MLM pretraining results at the base scale:
	GLUE dev \textit{performances (the higher the better)} in the upper panel and text classification \textit{error rates (the lower the better)} in the lower panel . 
	The FLOPs and \#Params both refer to the finetuning setting with only the encoder. 
	The corresponding numbers with the decoder are included in Appendix \ref{sec:appendix-training-cost}. 
	The FLOPs is a rough estimation assuming linear complexity w.r.t. the sequence length.
	The \#Params is exact including the embedding matrix.}
\label{tab:fair-comparison-mlm}
\vspace{-1.5em}
\end{table}
Here, we can make a few key observations:
\begin{itemize}[leftmargin=*,itemsep=0em,parsep=0em,topsep=0em]
\item Given similar or fewer FLOPs, by trading sequential resolution for more layers, the F-TFM outperforms the standard Transformer in most tasks except STS-B, especially for smaller models.
\item When we only compress the sequence length without increasing the depth (and \#Params), F-TFM could suffer from some performance loss in certain settings on the GLUE datasets.
However, as the model size increases, such performance gaps become smaller or even disappear.
\item In addition, we find partial parameter-sharing often harms the performance.
Therefore, the practical trade-off should be made according to the actual task and computation device.
\end{itemize}

To further test generality of F-TFM, we additionally consider ELECTRA for pretraining.
The results are summarized in Table~\ref{tab:fair-comparison-electra}.
Overall, we see a similar trend, though the gain is slightly smaller on the GLUE benchmark.
This could be attributed to reusing two key hyper-parameters (discriminator loss coefficient and generator size multiplier) tuned for Transformer to train F-TFMs without any adjustment at all. 
\begin{table}[!ht]
\centering
\small
\begin{tabular}{@{}lccccccccc@{}}
\toprule
Model size & CoLA & SST-2 & MRPC & STS-B & QQP & MNLI & QNLI & RTE & GLUE-AVG \\
\cmidrule(r){1-1} \cmidrule(lr){2-9} \cmidrule(l){10-10}
{L24H1024}
& 66.5 & 94.3 & 92.8/90.0 & 91.5 & 89.6/92.2 & 89.4 & 94.1 & 84.5 & 87.8 \\
{B10-10-10} 
& 68.6 & 95.0 & 93.0/90.0 & 91.0 & 88.9/91.7 & 89.1 & 93.6 & 84.5 & \bf 87.9 \\
{B8-8-8}
& 66.6 & 94.8 & 92.6/89.7 & 90.7 & 88.8/91.7 & 89.0 & 93.6 & 82.1 & 87.3 \\
\cmidrule(r){1-1} \cmidrule(lr){2-9} \cmidrule(l){10-10}
{L12H768}
& 64.3 & 93.1 & 92.1/89.2 & 90.8 & 88.7/91.7 & 86.4 & 92.1 & 75.4 & 85.4 \\
{B6-6-6} 
& 64.3 & 94.2 & 92.8/89.7 & 90.1 & 88.7/91.6 & 87.4 & 92.5 & 78.3 & \bf  86.0 \\
{B6-3x2-3x2}
& 63.9 & 94.2 & 93.0/90.2 & 89.5 & 88.4/91.4 & 87.0 & 92.2 & 77.6 & 85.7 \\
{B4-4-4}
& 62.8 & 93.6 & 92.5/89.2 & 89.2 & 88.4/91.3 & 86.0 & 91.6 & 74.3 & 84.8 \\
\cmidrule(r){1-1} \cmidrule(lr){2-9} \cmidrule(l){10-10}
{L6H768}
& 62.1 & 91.1 & 90.8/86.8 & 88.9 & 88.2/91.3 & 83.9 & 89.7 & 66.7 & 82.6 \\
{B3-4-4}
& 59.0 & 93.1 & 90.8/87.5 & 88.7 & 88.1/91.0 & 85.8 & 91.1 & 72.5 & \bf 83.6 \\
\bottomrule
\end{tabular}

\vspace{0.1em}

\begin{tabular}{@{}lccccccccc@{}}
\toprule
Model size & IMDB & AG & DBpedia & Yelp2 & Yelp5 & Amazon2 & Amazon5 & FLOPs & \#Params \\
\cmidrule(r){1-1} \cmidrule(lr){2-8} \cmidrule(l){9-10}
{L24H1024} 
& 4.724 & \bf 5.053 & 0.653 & 1.874 & 28.84 & 2.425 & 32.85 & {1.00x} & {1.00x} \\
{B10-10-10} 
& \bf 4.324 & 5.250 & \bf 0.639 & \bf 1.789 & \bf 28.68 & \bf 2.419 & \bf 32.72 & {0.73x} & {1.22x} \\
{B8-8-8}
& 4.364 & 5.408 & 0.651 & 1.729 & 28.76 & 2.447 & 32.85 & {0.58x} & {1.00x} \\
\cmidrule(r){1-1} \cmidrule(lr){2-8} \cmidrule(l){9-10}
{L12H768}
& 5.248 & 5.355 & 0.657 & 1.953 & 29.24 & 2.596 & 33.04 & {1.00x} & {1.00x} \\
{B6-6-6}
& \bf 4.792 & \bf 5.237 & \bf 0.650 & \bf 1.850 & \bf 28.73 & \bf 2.499 & \bf 32.79 & {0.88x} & {1.39x} \\
{B6-3x2-3x2}
& 4.924 & 5.342 & 0.671 & 1.913 & 29.00 & 2.523 & 32.85 & {0.88x} & {1.00x} \\
{B4-4-4}
& 5.152 & 5.382 & 0.659 & 2.032 & 29.33 & 2.566 & 33.03 & {0.58x} & {1.00x} \\
\cmidrule(r){1-1} \cmidrule(lr){2-8} \cmidrule(l){9-10}
{L6H768}
& 6.220 & 5.395 & 0.674 & 2.287 & 30.16 & 2.759 & 33.57 & {1.00x} & {1.00x} \\
{B3-4-4}
& \bf 5.396 & \bf 5.342 & \bf 0.653 & \bf 2.000 & \bf 29.60 & \bf 2.591 & \bf 33.09 & {1.00x} & {1.53x} \\
\bottomrule
\end{tabular}
\caption{ELECTRA pretraining results at the base scale.}
\label{tab:fair-comparison-electra}
\vspace{-1.5em}
\end{table}

\textbf{Running Time Comparison}\; While FLOPs count offers a general idea of the model speed, it still differs from the actual running time, especially when other overhead exists.
Hence, for completeness, we show the speedup provided by the F-TFM in terms of actual running time in Appendix~\ref{sec:appendix-training-cost}. We also compare the actual memory footprint of F-TFM and TFM in Appendix~\ref{sec:appendix-training-cost}.

\subsection{Large-scale Results}
\label{sec:exp-large-scale}
Given the encouraging results of F-TFM at base-scale, we next consider training F-TFM under the large-scale setting and compare it with previous models pretrained in similar settings.
Due to the slightly better performance of ELECTRA over MLM, we will use the ELECTRA objective for all large-scale experiments.
\begin{table}[!ht]
	\centering
	\small
	\begin{tabular}{l|c@{\hskip 0.05in}c@{\hskip 0.05in}c@{\hskip 0.05in}c@{\hskip 0.05in}c@{\hskip 0.05in}c@{\hskip 0.05in}c@{\hskip 0.05in}c@{\hskip 0.05in}c@{\hskip 0.05in}|c}
		\toprule
		\bf Model & \bf CoLA & \bf SST-2 & \bf MRPC & \bf STS-B & \bf QQP & \bf MNLI & \bf QNLI & \bf RTE & \bf WNLI & \bf AVG  \\
		\midrule
		\multicolumn{11}{c}{\textit{Dev set results (single model)}} \\
		\midrule
		ROBERTA$_\text{Large}$~\cite{liu2019roberta}
		& 68.0 & 96.4 &   -/90.9  & 92.4 & -/92.2    & 90.2 & 94.7 & 86.6 & - & 88.9 \\
		XLNet$_\text{Large}$~\cite{yang2019xlnet}
		& 69.0 & 97.0 &   -/90.8  & 92.5 & -/92.3    & 90.8 & 94.9 & 85.9 & - & 89.2 \\
		ELECTRA$_\text{Large}$~\cite{clark2020electra}
		& 69.1 & 96.9 &   -/90.8  & 92.6 & -/92.4    & 90.9 & 95.0 & 88.0 & - & 89.5 \\
		B10-10-10H1024
		& 72.4 & 96.8 & 93.5/90.9 & 92.1 & 89.8/92.4 & 91.1/- & 95.1 & 89.5 & - & \bf 90.0 \\
		B8-8-8H1024
		& 71.3 & 96.8 & 93.1/90.7 & 91.7 & 89.8/92.4 & 90.8/- & 94.7 & 89.2 & - & 89.7 \\
		\cmidrule(r){1-1} \cmidrule(lr){2-10} \cmidrule(l){11-11}
		ROBERTA$_\text{Base}$~\cite{liu2019roberta}
		& 63.6 & 94.8 & -/90.2    & 91.2 & -/91.9    & 87.6/- & 92.8 & 78.7 & - & 86.4 \\
		MPNet$_\text{Base}$~\cite{song2020mpnet}
		& 65.0 & 95.4 & -/91.5    & 90.9 & -/91.9    & 88.5/- & 93.3 & 85.2 & - & 87.7 \\
		B6-6-6H768
		& 70.1 & 96.3 & 93.2/90.4 & 91.1 & 89.2/92.0 & 89.7/- & 93.7 & 83.4 & - & \bf 88.3 \\
		B6-3x2-3x2H768 
		& 68.5 & 95.6 & 92.5/89.5 & 91.0 & 89.3/92.0 & 89.1/- & 93.0 & 83.4 & - & 87.8 \\
		B4-4-4H768
		& 68.2 & 95.0 & 92.8/90.2 & 90.3 & 89.0/91.8 & 88.6/- & 92.6 & 79.1 & - & 87.0 \\
		\midrule
		\multicolumn{11}{c}{\textit{Leaderboard test set results (single task \& single model)}} \\
		\midrule
		ELECTRA$_\text{Large}$~\cite{clark2020electra}
		& 68.1 & 96.7 & 89.2/92.0 & 92.1/91.7 & 74.8/90.4 & 90.7/90.2 & 95.5 & 86.1 & 65.1 & 85.2 \\
		B10-10-10H1024 
		& 68.9 & 97.2 & 89.4/92.1 & 91.6/91.3 & 74.3/90.2 & 90.9/90.9 & 95.5 & 86.5 & 65.1 & \bf 85.4 \\
		B8-8-8H1024
		& 68.3 & 96.9 & 89.2/92.0 & 91.5/91.1 & 73.8/90.1 & 90.7/90.7 & 95.1 & 85.3 & 65.1 & 85.0 \\
		\cmidrule(r){1-1} \cmidrule(lr){2-10} \cmidrule(l){11-11}
		ELECTRA$_\text{Base}$~\cite{clark2020electra}
		& 64.6 & 96.0 & 88.1/91.2 & 91.0/90.2 & 73.2/89.5 & 88.5/88.0 & 93.1 & 75.2 & 65.1 & 82.7  \\
		B6-6-6H768
		& 68.3 & 96.5 & 89.1/91.9 & 90.6/89.9 & 73.3/89.9 & 89.7/89.4 & 94.0 & 80.4 & 65.1 & \bf 84.0 \\
		B6-3x2-3x2H768
		& 65.9 & 96.0 & 87.8/91.0 & 90.0/89.6 & 73.3/89.8 & 88.9/88.7 & 93.8 & 79.9 & 65.1 & 83.4 \\
		\midrule
		\multicolumn{11}{c}{\textit{Leaderboard test set results (multi-task \& ensemble)}} \\
		\midrule
		ROBERTA$_\text{Large}$~\cite{liu2019roberta}
		& 67.8 & 96.7 & 89.8/92.3 & 92.2/91.9 & 74.3/90.2 & 90.8/90.2 & 95.4 & 88.2 & 89.0 & 88.1 \\
		ELECTRA$_\text{Large}$~\cite{clark2020electra}
		& 71.7 & 97.1 & 90.7/93.1 & 92.9/92.5 & 75.6/90.8 & 91.3/90.8 & 95.8 & 89.8 & 91.8 & 89.4 \\
		B10-10-10H1024
		& 70.5 & 97.5 & 91.2/93.4 & 92.6/92.3 & 75.4/90.7 & 91.4/91.1 & 95.8 & 90.0 & 94.5 & \bf 89.7 \\
		\bottomrule 
	\end{tabular}
	\caption{Comparison with previous methods on the GLUE benchmark under large-scale pretraining.}
	\label{tab:large-scale-glue}
	\vspace{-1em}
\end{table}
Given the pretrained F-TFM of different sizes, we first compare the finetuning performance on the GLUE benchmark in Table~\ref{tab:large-scale-glue}.
Similar to the base-scale results, with fewer or comparable FLOPs, F-TFM outperforms the corresponding baselines in the majority of tasks, suggesting the good scalability of F-TFM.
We also test the models on the 7 text classification tasks.
But due to the page constraint, we refer readers to Appendix~\ref{sec:appendix-textcls}.

Next, we consider the RACE dataset, which is quite different from the GLUE benchmark.
At the core, RACE is a multiple-choice reading comprehension task requiring complex reasoning, which though, can be formulated as classifying the correct choice.
Also, paragraphs in RACE are much longer.
To F-TFM, this presents both a challenge, as it requires detailed reasoning, and an opportunity to compress long paragraph.
As we can see in Table~\ref{tab:race}, F-TFM achieves better performances compared to all previous models.
In particular, within the base model group, the gain is very significant.
It shows that F-TFM can also excel for sequence-level task that involves long text and reasoning.

\begin{table}[!ht]
\centering
\small
\begin{minipage}{0.48\textwidth}
	\centering
	\begin{tabular}{lccc}
		\toprule
		\multirow{2}{*}{\bf Model} & \multicolumn{3}{c}{\bf RACE} \\
		& Total & High & Middle \\
		\cmidrule(r){1-1} \cmidrule(l){2-4}
		ROBERTA$_\text{Large}$~\cite{liu2019roberta}
		& 83.2 & 81.3 & 86.5 \\
		XLNet$_\text{Large}$~\cite{yang2019xlnet}
		& 85.4 & 84.0 & 88.6 \\
		B10-10-10
		& \bf 85.7 & \bf 84.4 & \bf 88.8 \\
		B8-8-8
		& 85.2 & 83.9 & 88.4 \\
		\cmidrule(r){1-1} \cmidrule(l){2-4}
		ALBERT$_\text{Base}$~\cite{lan2019albert}
		& 66.0 & -    & - \\
		MPNet$_\text{Base}$~\cite{song2020mpnet}
		& 72.0 & 76.3 & 70.3 \\
		B6-6-6
		& \bf 79.7 & \bf 78.2 & \bf 83.4 \\
		B6-3x2-3x2
		& 78.8 & 77.5 & 82.0 \\
		B4-4-4
		& 76.2 & 74.6 & 80.0 \\  
		\bottomrule 
	\end{tabular}
	\caption{RACE test performance comparison.}
	\label{tab:race}
\end{minipage}\hfill
\begin{minipage}{0.48\textwidth}
	\begin{tabular}{lcccc}
		\toprule
		\multirow{2}{*}{\bf Model} & \multicolumn{2}{c}{\bf SQuAD2.0} & \multicolumn{2}{c}{\bf SQuAD1.1} \\
		& EM & F1 & EM & F1 \\
		\cmidrule(r){1-1} \cmidrule(l){2-3} \cmidrule(l){4-5}
		ROBERTA$_\text{Large}$~\cite{liu2019roberta}
		& 86.5 & 89.4 & 88.9 & 94.6 \\
		ELECTRA$_\text{Large}$~\cite{clark2020electra}
		& \bf 88.0 & \bf 90.6 & \bf 89.7 & \bf 94.9 \\
		B10-10-10
		& 87.6 & 90.4 & 89.0 & 94.7 \\
		B8-8-8
		& 87.1 & 89.8 & 88.7 & 94.4 \\
		\cmidrule(r){1-1} \cmidrule(l){2-3} \cmidrule(l){4-5}
		ROBERTA$_\text{Base}$~\cite{liu2019roberta}
		& 80.5 & 83.7 & 84.6 & 91.5 \\
		MPNet$_\text{Base}$~\cite{lin2015hierarchical}
		& 80.5 & 83.3 & 86.8 & 92.5\\
		B6-6-6
		& \bf 85.1 & \bf 87.7 & \bf 87.4 & \bf 93.3 \\
		B6-3x2-3x2
		& 84.2 & 87.0 & 87.0 & 93.0 \\
		B4-4-4
		& 82.6 & 85.5 & 85.9 & 92.2 \\
		\bottomrule  
	\end{tabular}
	\caption{SQuAD dev performance comparison.}
	\label{tab:squad}
\end{minipage}
\vspace{-2em}
\end{table}
Finally, although F-TFM is mainly designed for tasks that only require a sequence-level representation, it is possible to apply F-TFM to token-level tasks by additionally finetuning the decoder.
To test this ability, we finetune F-TFM on the SQuAD datasets and compare it with previous models in Table~\ref{tab:squad}.
While F-TFM outperforms previous models in the base group by a large margin, in the large model group, the F-TFM with about 83\% FLOPs (B10-10-10) still falls behind the standard Transformer that always maintains a full-length token-level representations.
This suggests sequential compression could harm the performance when detailed token-level information is critical.
On the other hand, compared to the results on SQuAD1.1, F-TFMs perform relatively better on SQuAD2.0, which additionally requires the model to make a sequence-level prediction on whether the question is answerable.
This again shows the general effectiveness of the F-TFM in sequence-level tasks.

\subsection{Ablation Study}
\label{sec:ablation}
\begin{table}[!ht]
	\small
	\centering
	\begin{tabular}{@{}c|l@{\hskip 0.05in}lcccc|c@{}}
		\toprule
		ID & Layout &(FLOPs / Params) & Pool-Op & Pool-query-only & Sep \cls & Rel-Attn & GLUE-AVG  \\
		\midrule
		(1)& B6-6-6 & (1.00x / 1.00x)
		& Mean     & \checkmark & \checkmark & \checkmark & \bf 83.5 \\
		(2)&&& Mean     & \checkmark &            & \checkmark & 82.9 \\
		(3)&&& Mean     &            & \checkmark & \checkmark & 83.0 \\
		(4)&&& Mean     & \checkmark & \checkmark &            & 81.4 \\
		(5)&&& Max      & \checkmark & \checkmark & \checkmark & 83.4 \\
		(6)&&& Top-Attn & \checkmark & \checkmark & \checkmark & 75.8 \\
		(7)& B8-8  & (1.14x / 0.91x)
		& Mean     & \checkmark & \checkmark & \checkmark & 83.4 \\
		(8)& B5-5-5-5 & (0.89x / 1.08x)
		& Mean     & \checkmark & \checkmark & \checkmark & 82.9 \\
		\bottomrule
	\end{tabular}
	\caption{Ablation study of F-TFMs with different designs.}
	\label{tab:ablation}
\end{table}

Finally, based on the GLUE benchmark, we perform a series of ablation studies on the importance of various designs in F-TFM, including the block layout design, the type of pooling operation, the pool-query-only technique, maintaining a separate \cls vector and the usage of Transformer-XL parameterization.
\begin{itemize}[leftmargin=*]
	\item Pooling operation: Including the mean pooling we finally employ in F-TFM, we actually test two types of pooling operations.
	\begin{itemize}
		\item[(1)] The first type is just the strided mean/max pooling as described in section~\ref{sec:method}.
		\item[(2)] The second type aims to select a subset of ``hub'' states, which refer to those hidden vectors that are attended most in the previous S-Attn layer and hence likely to carry most critical information about the sequence.
		Concretely, given the attention map from the previous S-Attn layer, we reduce sum the scores along the number of head and query length dimensions to a score for each position.
		Then, we simply choose the top 50\% of states to achieve the same compression rate.
		Note that, this type of pooling operation is essentially the same as the important states selection procedure in Power-BERT~\cite{goyal2020power}.
	\end{itemize}
	\item Pool-query-only design
	\item Separating \cls in the pooling operation
	\item Block layout design: In our experiments, all models actually utilize a 3-block design. Here, we compare the 3-blocks design with the 2-blocks and the 4-blocks design. 
	\item Relative attention parameterization proposed in Transformer-XL \cite{dai2019transformer}.
	We compare this parameterization with the learned absolute position embedding as used in the BERT~\cite{devlin2018bert}.
\end{itemize}

The ablation results are included in Table~\ref{tab:ablation}. 
To save the computation resources, the size of model hidden states in table \ref{tab:ablation} is set as 512. 
From the ablation results, we can make the following observations:
\begin{itemize}[leftmargin=*]
	\item Comparing pooling different operation ((1), (5), and (6)), we found that the performance of the mean and max pooling operation is similar.
	But they are significantly better than the idea of utilizing attention score (Top-Attn pooling) to select the ``hub'' states.
	
	\item Comparing (1) with (2) and (3) respectively, we see that the two special designs, i.e. ``pool-query-only'' and maintaining a separate non-pooled \cls, can both bring a clear improvement to the proposed model.
	
	\item Comparing (1) and (4), we find that the relative positional parameterization is key to the performance of the proposed F-TFM.
	We suspect that the pooling operation could destroy the positional information carried by the absolute position encoding, which is only injected to the model in the input embedding layer.
	As a result, the higher blocks may not have enough positional information to learn a good enough attention pattern.
	In comparison, the positional information is injected to each layer under the relative positional attention scheme.
	Therefore, to achieve good result with F-TFM based on absolute positional embedding, one may inject the absolute positional embedding into each attention layer.
	Actually, a contemporary application of Transformer to the detection problem in computer vision shows injecting positional embedding into each layer is important~\cite{Carion2020EndtoEndOD}.
	
	\item Finally, we study the influence of block layout design in our framework.
	With B6-6-6 as the 3-block benchmark, we consider two other layout design with similar FLOPs and number of parameters. Specifically, we consider B8-8 for the 2-block design and B5-5-5-5 for the 4-block design.
	Comparing the results in (1), (7), and (8), we find that the performance of the 3-block (B6-6-6) design achieves the best performance, which is significantly better than the 4-block design and slightly better than the 2-block design.
	However, if we further taking the FLOPs/\#Params into consideration, it is more clear that the 3-block design is superior.
	Therefore, in the main paper, we always use the 3-block design.
\end{itemize}

\section{Conclusion \& Discussion}
\label{sec:summary}
In this work, under the pretraining-finetuning paradigm, we investigate a largely overlooked dimension of complexity in language processing.
With the proposed Funnel-Transformer, we show how sequential resolution can be compressed in a simple form to save computation and how the saved FLOPs can be re-invested in improving the model capacity and hence the performance.
Open challenges for future research include the better ways to improve the compression scheme, to optimize the block layout design and to re-invest the saved FLOPs.
In addition, combining Funnel-Transformer with model compression techniques like knowledge distillation and quantization would be an important direction towards the enhancement of practical impact.


\bibliographystyle{unsrt}
\bibliography{ref}

\begin{thebibliography}{10}

\bibitem{peters2018deep}
Matthew~E Peters, Mark Neumann, Mohit Iyyer, Matt Gardner, Christopher Clark,
  Kenton Lee, and Luke Zettlemoyer.
\newblock Deep contextualized word representations.
\newblock {\em arXiv preprint arXiv:1802.05365}, 2018.

\bibitem{devlin2018bert}
Jacob Devlin, Ming-Wei Chang, Kenton Lee, and Kristina Toutanova.
\newblock Bert: Pre-training of deep bidirectional transformers for language
  understanding.
\newblock {\em arXiv preprint arXiv:1810.04805}, 2018.

\bibitem{yang2019xlnet}
Zhilin Yang, Zihang Dai, Yiming Yang, Jaime Carbonell, Russ~R Salakhutdinov,
  and Quoc~V Le.
\newblock Xlnet: Generalized autoregressive pretraining for language
  understanding.
\newblock In {\em Advances in neural information processing systems}, pages
  5754--5764, 2019.

\bibitem{liu2019roberta}
Yinhan Liu, Myle Ott, Naman Goyal, Jingfei Du, Mandar Joshi, Danqi Chen, Omer
  Levy, Mike Lewis, Luke Zettlemoyer, and Veselin Stoyanov.
\newblock Roberta: A robustly optimized bert pretraining approach.
\newblock {\em arXiv preprint arXiv:1907.11692}, 2019.

\bibitem{clark2020electra}
Kevin Clark, Minh-Thang Luong, Quoc~V Le, and Christopher~D Manning.
\newblock Electra: Pre-training text encoders as discriminators rather than
  generators.
\newblock {\em arXiv preprint arXiv:2003.10555}, 2020.

\bibitem{lan2019albert}
Zhenzhong Lan, Mingda Chen, Sebastian Goodman, Kevin Gimpel, Piyush Sharma, and
  Radu Soricut.
\newblock Albert: A lite bert for self-supervised learning of language
  representations.
\newblock {\em arXiv preprint arXiv:1909.11942}, 2019.

\bibitem{kong2019mutual}
Lingpeng Kong, Cyprien de~Masson d'Autume, Wang Ling, Lei Yu, Zihang Dai, and
  Dani Yogatama.
\newblock A mutual information maximization perspective of language
  representation learning.
\newblock {\em arXiv preprint arXiv:1910.08350}, 2019.

\bibitem{raffel2019exploring}
Colin Raffel, Noam Shazeer, Adam Roberts, Katherine Lee, Sharan Narang, Michael
  Matena, Yanqi Zhou, Wei Li, and Peter~J Liu.
\newblock Exploring the limits of transfer learning with a unified text-to-text
  transformer.
\newblock {\em arXiv preprint arXiv:1910.10683}, 2019.

\bibitem{lewis2019bart}
Mike Lewis, Yinhan Liu, Naman Goyal, Marjan Ghazvininejad, Abdelrahman Mohamed,
  Omer Levy, Ves Stoyanov, and Luke Zettlemoyer.
\newblock Bart: Denoising sequence-to-sequence pre-training for natural
  language generation, translation, and comprehension.
\newblock {\em arXiv preprint arXiv:1910.13461}, 2019.

\bibitem{song2019mass}
Kaitao Song, Xu~Tan, Tao Qin, Jianfeng Lu, and Tie-Yan Liu.
\newblock Mass: Masked sequence to sequence pre-training for language
  generation.
\newblock {\em arXiv preprint arXiv:1905.02450}, 2019.

\bibitem{liu2019multi}
Xiaodong Liu, Pengcheng He, Weizhu Chen, and Jianfeng Gao.
\newblock Multi-task deep neural networks for natural language understanding.
\newblock {\em arXiv preprint arXiv:1901.11504}, 2019.

\bibitem{song2020mpnet}
Kaitao Song, Xu~Tan, Tao Qin, Jianfeng Lu, and Tie-Yan Liu.
\newblock Mpnet: Masked and permuted pre-training for language understanding.
\newblock {\em arXiv preprint arXiv:2004.09297}, 2020.

\bibitem{vaswani2017attention}
Ashish Vaswani, Noam Shazeer, Niki Parmar, Jakob Uszkoreit, Llion Jones,
  Aidan~N Gomez, {\L}ukasz Kaiser, and Illia Polosukhin.
\newblock Attention is all you need.
\newblock In {\em Advances in neural information processing systems}, pages
  5998--6008, 2017.

\bibitem{so2019evolved}
David~R So, Chen Liang, and Quoc~V Le.
\newblock The evolved transformer.
\newblock {\em arXiv preprint arXiv:1901.11117}, 2019.

\bibitem{chen2020adabert}
Daoyuan Chen, Yaliang Li, Minghui Qiu, Zhen Wang, Bofang Li, Bolin Ding, Hongbo
  Deng, Jun Huang, Wei Lin, and Jingren Zhou.
\newblock Adabert: Task-adaptive bert compression with differentiable neural
  architecture search.
\newblock {\em arXiv preprint arXiv:2001.04246}, 2020.

\bibitem{wu2020lite}
Zhanghao Wu, Zhijian Liu, Ji~Lin, Yujun Lin, and Song Han.
\newblock Lite transformer with long-short range attention.
\newblock {\em arXiv preprint arXiv:2004.11886}, 2020.

\bibitem{wu2019pay}
Felix Wu, Angela Fan, Alexei Baevski, Yann~N Dauphin, and Michael Auli.
\newblock Pay less attention with lightweight and dynamic convolutions.
\newblock {\em arXiv preprint arXiv:1901.10430}, 2019.

\bibitem{sun2020mobilebert}
Zhiqing Sun, Hongkun Yu, Xiaodan Song, Renjie Liu, Yiming Yang, and Denny Zhou.
\newblock Mobilebert: a compact task-agnostic bert for resource-limited
  devices.
\newblock {\em arXiv preprint arXiv:2004.02984}, 2020.

\bibitem{subramanian2020multi}
Sandeep Subramanian, Ronan Collobert, Marc'Aurelio Ranzato, and Y-Lan Boureau.
\newblock Multi-scale transformer language models.
\newblock {\em arXiv preprint arXiv:2005.00581}, 2020.

\bibitem{goyal2020power}
Saurabh Goyal, Anamitra~Roy Choudhary, Venkatesan Chakaravarthy, Saurabh
  ManishRaje, Yogish Sabharwal, and Ashish Verma.
\newblock Power-bert: Accelerating bert inference for classification tasks.
\newblock {\em arXiv preprint arXiv:2001.08950}, 2020.

\bibitem{lin2015hierarchical}
Rui Lin, Shujie Liu, Muyun Yang, Mu~Li, Ming Zhou, and Sheng Li.
\newblock Hierarchical recurrent neural network for document modeling.
\newblock In {\em Proceedings of the 2015 Conference on Empirical Methods in
  Natural Language Processing}, pages 899--907, 2015.

\bibitem{zhang2019hibert}
Xingxing Zhang, Furu Wei, and Ming Zhou.
\newblock Hibert: Document level pre-training of hierarchical bidirectional
  transformers for document summarization.
\newblock {\em arXiv preprint arXiv:1905.06566}, 2019.

\bibitem{garg2019multiresolution}
Vikas~K Garg, Inderjit~S Dhillon, and Hsiang-Fu Yu.
\newblock Multiresolution transformer networks: Recurrence is not essential for
  modeling hierarchical structure.
\newblock {\em arXiv preprint arXiv:1908.10408}, 2019.

\bibitem{ronneberger2015u}
Olaf Ronneberger, Philipp Fischer, and Thomas Brox.
\newblock U-net: Convolutional networks for biomedical image segmentation.
\newblock In {\em International Conference on Medical image computing and
  computer-assisted intervention}, pages 234--241. Springer, 2015.

\bibitem{scherer2010evaluation}
Dominik Scherer, Andreas M{\"u}ller, and Sven Behnke.
\newblock Evaluation of pooling operations in convolutional architectures for
  object recognition.
\newblock In {\em International conference on artificial neural networks},
  pages 92--101. Springer, 2010.

\bibitem{ying2018hierarchical}
Zhitao Ying, Jiaxuan You, Christopher Morris, Xiang Ren, Will Hamilton, and
  Jure Leskovec.
\newblock Hierarchical graph representation learning with differentiable
  pooling.
\newblock In {\em Advances in neural information processing systems}, pages
  4800--4810, 2018.

\bibitem{gao2019graph}
Hongyang Gao and Shuiwang Ji.
\newblock Graph u-nets.
\newblock {\em arXiv preprint arXiv:1905.05178}, 2019.

\bibitem{lee2019self}
Junhyun Lee, Inyeop Lee, and Jaewoo Kang.
\newblock Self-attention graph pooling.
\newblock {\em arXiv preprint arXiv:1904.08082}, 2019.

\bibitem{wang2018glue}
Alex Wang, Amanpreet Singh, Julian Michael, Felix Hill, Omer Levy, and Samuel~R
  Bowman.
\newblock Glue: A multi-task benchmark and analysis platform for natural
  language understanding.
\newblock {\em arXiv preprint arXiv:1804.07461}, 2018.

\bibitem{zhang2015character}
Xiang Zhang, Junbo Zhao, and Yann LeCun.
\newblock Character-level convolutional networks for text classification.
\newblock In {\em Advances in neural information processing systems}, pages
  649--657, 2015.

\bibitem{lai2017race}
Guokun Lai, Qizhe Xie, Hanxiao Liu, Yiming Yang, and Eduard Hovy.
\newblock Race: Large-scale reading comprehension dataset from examinations.
\newblock {\em arXiv preprint arXiv:1704.04683}, 2017.

\bibitem{dai2019transformer}
Zihang Dai, Zhilin Yang, Yiming Yang, Jaime Carbonell, Quoc~V Le, and Ruslan
  Salakhutdinov.
\newblock Transformer-xl: Attentive language models beyond a fixed-length
  context.
\newblock {\em arXiv preprint arXiv:1901.02860}, 2019.

\bibitem{Carion2020EndtoEndOD}
Nicolas Carion, F.~Massa, Gabriel Synnaeve, Nicolas Usunier, Alexander~M
  Kirillov, and Sergey Zagoruyko.
\newblock End-to-end object detection with transformers.
\newblock {\em ArXiv}, abs/2005.12872, 2020.

\bibitem{xie2019unsupervised}
Qizhe Xie, Zihang Dai, Eduard Hovy, Minh-Thang Luong, and Quoc~V Le.
\newblock Unsupervised data augmentation for consistency training.
\newblock 2019.

\end{thebibliography}

\clearpage
\appendix
\section{Implementation Optimization}
\subsection{Sequence Truncation for Separating \cls trick}
\label{sec:appendix-seq-trunc}
As discussed in Section \ref{sec:proposed}, to avoid breaking the \cls structure commonly used in pretraining, we do not apply the pooling operation to the \cls and keep the hidden state corresponding to \cls intact.
While conceptually simple, a naive implementation could slow down the computation by 15\% due to the ``irregular'' sequence length caused by such an operation.
Specifically, assume that sequence length of an input sample is a power of two, i.e., $2^p$, which usually is 512 in the pretraining phase.
After one pooling operation with the \cls intact, the length of the pooled sequence becomes $2^{p-1} + 1$, which is not a power of 2 anymore.
As a result, it can cause memory misalignment and the waste of paralleled computation power in accelerators, leading to substantial speed loss.

To resolve this issue, we employ a simple strategy to truncate the last token after the pooling.
Formally, denoting the pooled hidden state as $\rvh = \{h_{\cls}, h_1, \cdots,  h_{2^{p-1}}\} $, the truncation can be expressed as
\begin{equation}
\hat{\rvh} = \text{truncate}(\rvh) = \seq{h_{\cls}, h_1, \cdots,  h_{2^{p-1}-1}}
\end{equation}
With this simple trick, we can always keep the sequence length a power of 2, hence avoiding the slowdown caused by maintaining an independent \cls hidden state.

\subsection{Relative Positional Attention Implementation}
\label{sec:appendix-rel-attention}
In this work, we use the relative positional attention parameterization proposed in the Transformer-XL~\cite{dai2019transformer}.
To facilitate further discussion, we first review the details of this parameterization.
Taking the case of single head attention as the example head. 
Let $T, D$ be the sequence length and hidden dimension respectively.
Then, the pre-softmax attention score $A_{ij}$ between a pair of positions $i$ and $j$ consists of two terms:
\begin{align}
\label{eqn:rel_attn}
A_{ij} &= 
	\underbrace{ (W_Q h_i + v)^\top (W_K h_j) }_\text{ content term } + \underbrace{ (W_Q h_i + u)^\top (W_R r_{i-j}) }_\text{ position term }.
\end{align}
where $v, u \in \R^{D}$ are two trainable bias vectors, $W_Q, W_K, W_R \in \R^{D \times D}$ are three trainable projection matrices, and $r_{i-j} \in \R^{D}$ is the sinusoidal positional encoding that represents the relative distance $i-j$ between the two positions.

To compute the entire attention score matrix $\mb{A}$, the content term can easily be obtained via two head projections and an \textit{outer product} of complexity $O(TD^2 + T^2D)$:
\[ 
	\mb{A}^\text{content} = (\mb{H}W_Q + v) (\mb{H}W_K)^\top,
\]
where $\mb{H} = \seq{h_1, \cdots, h_T} \in \R^{T \times D}$ collects all hidden states into a matrix.
However, we cannot compute the position term in the same way as each $A_{ij}^\text{position}$ corresponds to a different $r_{i-j}$. 
Hence, a naive solution will be stacking $T^2$ pairs of position encodings into a tensor $\mb{\hat{R}} \in \R^{T\times T\times D}$ where $\mb{\hat{R}}_{ij} = r_{i-j}$, and then perform the following tensor product:
\[
	\mb{A}^\text{position} = \texttt{einsum}(\texttt{"id,ijd->ij"}, \mb{H}W_Q + u, \hat{\mb{R}}W_R).
\]
Note that the head projection $\mb{R}W_K$ now has a complexity of $O(T^2D^2)$ and a memory footprint of $O(T^2D)$, dominating all other computations.

\subsubsection{Standard Solution: Gather / Shift}
To resolve the computation burden above, a common technique is to instead collect a matrix $\mb{R} \in \R^{2T-1 \times D}$, where 
\[ \mb{R} = \seq{r_{T-1}, \dots, r_{0}, \cdots, r_{1-T}} \]
which includes all possible position encodings arranged from the maximum possible distance value $T-1$ to the minimum one $1 - T$.
Note that the full $\mb{\hat{R}}$ can be formed by gathering specific elements from $\mb{R}$ with an index matrix $\mb{I}$ of shape $[T \times T]$, i.e., 
\[ \mb{\hat{R}} = \text{gather}(\mb{R}, \mb{I}), \quad I_{ij} = T + i - j. \]
Mathematically, this is equivalent to using a permutation tensor $\mb{P} \in \R^{T \times T \times 2T-1}$ to multiply $\mb{R}$, i.e., $\mb{\hat{R}} = \mb{P} \mb{R}$, where $\mb{P}_{ij} \in \R^{2T-1}$ is a one-hot vector used to select/gather a single position of $\mb{R}$.
As the attention score computation only involves linear operations, we can rearrange the computation of the position term as follows
\begin{align*}
\mb{A}^\text{position} 
&= \texttt{einsum}\rbr{ \texttt{"id,ijd->ij"}, \mb{H}W_Q + u, \green{(\mb{P}\mb{R})} W_R } \\
&= \texttt{einsum}\rbr[\Big]{ \texttt{"ijk,jk->ij"}, \green{\mb{P}}, \sbr{ (\mb{H}W_Q + v) (\mb{R}W_R)^\top } } \\
&= \texttt{gather}\rbr[\Big]{ (\mb{H}W_Q + v) (\mb{R}W_R)^\top, \mb{I} }
\end{align*}
Note that, assuming gathering $T^2$ elements only has a complexity of $O(T^2)$, which is true for CPU/GPU, this trick reduces the computation complexity back to $O(2TD^2 + 2T^2D)$.
In practice, the gather operation can be implemented via a smart reshape operation, that is even cheaper.

\subsubsection{Optimization for TPU: factorized relative positional attention}
However, on TPUs, the assumption that gathering $T^2$ elements only has a complexity of $O(T^2)$ does not hold. 
Instead, we found that such a gather operation is dramatically slower on TPU.
Hence, we here consider another implementation which is significantly faster on TPU.

Firstly, let's rewrite the position term as follows
\begin{align}
A^\text{position}_{ij} 
&=  (W_Q h_i + u)^\top (W_R r_{i-j}) \nonumber\\
&=  \sbr[\Big]{\; \underbrace{ W_R^\top \rbr{ W_Q h_i + u}  }_{q_i} \;}^\top r_{i-j} \nonumber\\
\label{eqn:factorized-0}
&= q_i^\top r_{i-j}.
\end{align}
For easier derivation, we have introduced a notation of $q_i$. 
Then, recall the $r_{i-j}$ is the sinusoidal encoding that consists of the sine and the cosine components $r_{i-j} = \texttt{cat}(\sin_{i-j}, \cos_{i-j})$, where
\begin{align*}
\sin_{t} &= \seq{ \sin\rbr{t / 10000^{2/D}}, \sin\rbr{t / 10000^{4/D}}, \cdots, \sin\rbr{t / 10000^{D/D}} } \in \R^{D/2}, \\
\cos_{t} &= \seq{ \cos\rbr{t / 10000^{2/D}}, \cos\rbr{t / 10000^{4/D}}, \cdots, \cos\rbr{t / 10000^{D/D}} } \in \R^{D/2}.
\end{align*}
Hence, we similarly divide $q_i$ defined above into two parts, i.e., 
\[ q_i = \texttt{cat}(q_i^{\sin}, q_i^{\cos}). \]

Given the definitions, we can further break Eqn. \eqref{eqn:factorized-0} into two terms:
\begin{align*}
A^\text{position}_{ij} &= q_i^\top r_{i-j} = {q_i^{\sin}}^\top \sin_{i-j} + {q_i^{\cos}}^\top \cos_{i-j}.
\end{align*}
Now, using the trigonometric identities $\sin(a - b) = \sin(a)\cos(b) - \cos(a)\sin(b)$ and $\cos(a - b) = \cos(a)\cos(b) + \sin(a)\sin(b)$, the two terms can be respectively reformulated into
\begin{align*}
{q_i^{\sin}}^\top \sin_{i-j} 
&= {q_i^{\sin}}^\top \sbr{\sin_{i} \odot \cos_{j} - \cos_{i} \odot \sin_{j}} \\
&= {q_i^{\sin}}^\top \rbr{\sin_{i} \odot \cos_{j}} - {q_i^{\sin}}^\top \rbr{\cos_{i} \odot \sin_{j}} \\
&= \sbr{ q_i^{\sin} \odot \sin_{i} }^\top \cos_{j} + \sbr{ q_i^{\sin} \odot (-\cos_{i}) }^\top \sin_{j} \\
\intertext{and}
{q_i^{\cos}}^\top \cos_{i-j} 
&= {q_i^{\cos}}^\top \sbr{\cos_{i} \odot \cos_{j} + \sin_{i} \odot \sin_{j}} \\
&= {q_i^{\cos}}^\top \rbr{\cos_{i} \odot \cos_{j}} + {q_i^{\cos}}^\top \rbr{\sin_{i} \odot \sin_{j}} \\
&= \sbr{ q_i^{\cos} \odot \cos_{i} }^\top \cos_{j} + \sbr{ q_i^{\cos} \odot \sin_{i} }^\top \sin_{j}
\end{align*}

Hence, combining these two parts together, it follows that
\begin{align*}
q_i^\top r_{i-j} 
&= {q_i^{\sin}}^\top \sin_{i-j} + {q_i^{\cos}}^\top \cos_{i-j} \\
&= \sbr{ q_i^{\sin} \odot \sin_{i} }^\top \cos_{j} 
+ \sbr{ q_i^{\sin} \odot (-\cos_{i}) }^\top \sin_{j} 
+ \sbr{ q_i^{\cos} \odot \cos_{i} }^\top \cos_{j} 
+ \sbr{ q_i^{\cos} \odot \sin_{i} }^\top \sin_{j} \\
&= \cbr{ 
	\sbr{ q_i^{\sin} \odot \sin_{i} }^\top \cos_{j} + 
	\sbr{ q_i^{\cos} \odot \cos_{i} }^\top \cos_{j} } 
+ \cbr{
	\sbr{ q_i^{\sin} \odot (-\cos_{i}) }^\top \sin_{j} +
	\sbr{ q_i^{\cos} \odot \sin_{i} }^\top \sin_{j} } \\
&= \sbr[\bigg]{ 
		\underbrace{ \texttt{cat}( q_i^{\sin}, q_i^{\cos} ) }_{= q_i} \odot 
		\underbrace{ \texttt{cat}(\sin_{i}, \cos_{i}) }_{\coloneqq \phi_i} 
	}^\top \underbrace{ \texttt{cat}(\cos_{j}, \cos_{j} ) }_{\coloneqq \psi_j}\\ &\quad + 
   \sbr[\bigg]{ 
   		\underbrace{ \texttt{cat}(q_i^{\sin}, q_i^{\cos}) }_{= q_i} \odot 
   		\underbrace{ \texttt{cat}(-\cos_{i}, \sin_{i}) }_{\coloneqq \pi_i} 
   	}^\top \underbrace{ \texttt{cat}(\sin_{j}, \sin_{j})  }_{\coloneqq \omega_j} \\
&= \sbr{q_i \odot \phi_i }^\top \psi_j + \sbr{q_i \odot \pi_i}^\top \omega_j,
\end{align*}
where $\phi_i, \psi_j, \pi_i, \omega_j$ above are simply 4 positional encodings formed by concatenating the cosine and sine vectors of the corresponding $i$ and $j$ in different ways.
Note that, each term of the last line has a \textit{factorized} form that can be computed via an \textit{outer product}, just like the standard content term.
Therefore, by stacking $\phi_i, \psi_j, \pi_i, \omega_j$ of all positions (i.e. $i=1,\dots,T$ and $j=1,\dots,T$) into the corresponding $\mb{\Phi}, \mb{\Psi}, \mb{\Pi}, \mb{\Omega} \in \R^{T \times D}$ respectively, the full position term can be expressed in a simple form
\begin{align*}
\mb{A}^\text{position} 
&= \cbr[\Big]{ \sbr[\big]{ \rbr{ \mb{H} W_Q + u} W_R^\top } \odot \mb{\Phi}  } \mb{\Psi}^\top 
+ \cbr[\Big]{ \sbr[\big]{ \rbr{ \mb{H} W_Q + u} W_R^\top } \odot \mb{\Pi}  } \mb{\Omega}^\top 
\end{align*}
which leads to the complexity of $O(2TD^2 + 4T^2D)$, which is comparable to the content term.

\subsection{Potential Model Extensions}

In this section, we discuss some potential model extensions of Funnel-Transformer.
As described in section~\ref{sec:method}, \name can be divided into an encoder with a compression functionality and a decoder that recovers the full-length token-level representations.
To further extend the proposed model, first note that the encoder-decoder framework can be formulated into a more general form:
\begin{align*}
\rvh_\text{enc} &= \text{Encoder}(\rvx_\text{enc}), \\
\rvh_\text{dec} &= \text{Decoder}(\rvh_\text{enc}, \rvx_\text{dec}),
\end{align*}
where $\rvx_\text{enc}$ and $\rvx_\text{dec}$ are the encoder input sequence and the \textit{optional} and \textit{problem-specific} decoder input, respectively.
The goal of encoder is to compressing the input sequence $\rvx_\text{enc}$ into the hidden representations $\rvh_\text{enc}$ with a reduced length.
Then, conditioned on the decoder input $\rvh_\text{enc}$ if any, the decoder will extract relevant information/representations from $\rvh_\text{enc}$ to solve the specific NLP problem at hand.
Next, we will how the general form of \name can be instantiated into specific forms to solve corresponding NLP problems.

\paragraph{Sequence-level prediction} This is essentially the case we consider in most of our experiments where we want to obtain a vectorial representation of the input sequence such as text classification.
In this case, we don't really need the decoder $\rvx_\text{dec}$ (i.e. $\rvx_\text{dec} = \varnothing$) and the decoder simply extracts the hidden representation corresponding to the \cls token from $\rvh_\text{enc}$ and feeds it into the task-specific structure (e.g. classifier).

\paragraph{Token-level prediction} In the token-level prediction tasks such as the MLM pretraining, SQuAD and sequence labeling, we need a decoder to recover the token-level representations from the compressed sequence $\rvh_\text{enc}$.
In many cases, $\rvx_\text{dec}$ could simply be the original sequence or a token-level hidden representation of it to provide fine grained low-level information of each token and hence ease the optimization.
In this paper, we utilize the last-layer hidden states of the 1st block (before the first pooling operation) as the additional decoder input.

But for problems that utilize additional input signals, such as the permutation order used for permuted language modeling in XLNet~\cite{yang2019xlnet}.
This additional information can be injected into Funnel-Transformer via the decoder input $\rvx_\text{dec}$ to (approximately) recover some more complex control of attention mechanism.

\paragraph{Sequence-to-sequence problems} Another important category of NLP task is sequence-to-sequence problems, including machine translation, text summarization, and dialog generation, whose state-of-the-art solution is the conventional encoder-decoder framework.
Hence, \name naturally fits these tasks, where the decoder input $\text{x}_\text{dec}$ corresponds to the target text sequence and the encoder input $\text{x}_\text{enc}$ the source text sequence.
This way, the key difference compared to conventional models is the source side compression Funnel-Transformer provides.

Overall, we summarize some potential directions to extend \name presented in section \ref{sec:proposed} to NLP problems.
Finally, although we focus on discussion on the NLP tasks in this paper, \name  could be applied to any tasks dealing with sequential data, such as time series and video stream analysis.

\section{Experiment Setting and Hyper-parameters}
\label{sec:appendix-hparam}
\subsection{Preprocessing \& Tokenization}
For all experiments conducted in this work, we simply adapt the ``uncased'' word piece model originally used by BERT~\cite{devlin2018bert}, where the vocabulary size is about 30K.
Other than lower case and the default preprocessing included in the word piece tokenizer, the only additional preprocessing we perform is to remove some http symbols (e.g. \texttt{<b>}) in the 7 text classification tasks.

\subsection{Pretraining}
\begin{table}[h!]
	\centering
	\begin{tabular}{lcc}
		\toprule
		\bf Hparam          & \bf Base Scale & \bf Large Scale \\
		\midrule
		Hidden dropout      & \multicolumn{2}{c}{0.1} \\
		GeLU dropout        & \multicolumn{2}{c}{0.0} \\
		Attention dropout   & \multicolumn{2}{c}{0.1} \\
		Max sequence length & \multicolumn{2}{c}{512} \\
		Batch size          & 256  & 8192 \\
		Learning rate       & 1e-4 & 2e-4 \\
		Number of steps     & 1M   & 500K \\
		Warmup steps        & 10K  & 30K \\
		Optimizer 		    & \multicolumn{2}{c}{Adam Weight Decay} \\
		Learning rate decay & \multicolumn{2}{c}{Linear} \\
		Adam epsilon        & \multicolumn{2}{c}{1e-6} \\
		Weight decay        & \multicolumn{2}{c}{0.01} \\
		\bottomrule
	\end{tabular}
	\caption{Hyper-parameters for pretraining.}
	\label{tab:hp-pretrain}
	\vspace{-1em}
\end{table}

The hyper-parameters used for the two different pretraining settings are summarized in Table \ref{tab:hp-pretrain}.
One exception is the learning rate used for B10-10-10H1024 at the base scale. 
Specifically, we find the training can be unstable when the depth goes beyond 24 layers (in the case of B10-10-10H1024) at base scale, especially for the MLM objective.
Hence, we reduce the learning to 8e-5 for the B10-10-10H1024 F-TFM during base-scale pretraining.
This has a side effect of a slower training pace and potentially a slightly worse finetuning performance.
However, we does not observe such instability when the batch size is increased such as in the large-scale setting.

For ELECTRA, there are two additional important hyper-parameters, i.e., the discriminator loss coefficient and the relative size multiplier of the generator.
In this work, we does not tune these two hyper-parameters at all and simply use the numbers from the original paper, i.e., the discriminator loss coefficient of 50 and size multiplier of 1/4 for all architectures trained with ELECTRA.
In addition, in ELECTRA training, whenever F-TFM is used as the discriminator, the generator also uses the F-TFM.

In additional, in the all experiments, we only annotate the size of hidden states the rest of model sizes can be derived from on it:
\begin{itemize}
\item The embedding size = hidden size
\item The size of inner states of P-FFN is ``$4 \times \text{hidden size}$''.
\item The attention head dimension is always $64$.
\item The number of attention heads is ``$\text{hidden size} / 64$''.
\end{itemize}

Finally, another important element in pretraining is the mask sampling strategy.
For MLM training, following previous work, we always complete word span (up to 5 complete words) sampling.
However, for ELECTRA training, we notice a weird phenomenon that under the base-scale setting, the performance of both the Transformer and the F-TFM drops significantly if we use word span sampling rather than the single-token sampling.
On the other hand, under the large-scale setting, using word span sampling works fine.
Hence, we use single-token sampling for base-scale ELECTRA training, and word span sampling for large-scale ELECTRA training.

\subsection{Finetuning}

\begin{table}[h!]
\centering
\begin{tabular}{lcccccccc}
	\toprule
	\bf Hparam & \bf RTE & \bf MRPC	& \bf STS-B	& \bf CoLA & \bf SST-2 & \bf QNLI &\bf MNLI & \bf QQP \\
	\midrule
	Hidden dropout
	& \multicolumn{8}{c}{0.1} \\
	GeLU dropout
	& \multicolumn{8}{c}{0.0} \\
	Attention dropout
	& \multicolumn{8}{c}{0.1} \\
	Max sequence length
	& \multicolumn{8}{c}{128}  \\
	Batch size
	& 16 & 16 & 16 & 16 & 32 & 32 & 64 & 64 \\
	Number of epochs
	& 10 & 10 & 10 & 10 &  5 &  3 &  3 &  5 \\
	Learning rate decay & \multicolumn{8}{c}{Linear} \\
	Weight decay        & \multicolumn{8}{c}{0.01}   \\
	Warmup proportion   & \multicolumn{8}{c}{0.1}   \\
	Adam epsilon        & \multicolumn{8}{c}{1e-6}   \\
	\bottomrule
\end{tabular}
\begin{tabular}{lccccccc}
	\toprule
	\bf Hparam & \bf IMDB & \bf AG & \bf DBpedia & \bf Yelp-2 & \bf Yelp-5 & \bf Amazon-2 & \bf Amazon-5 \\
	\midrule
	Hidden dropout
	& \multicolumn{7}{c}{0.1} \\
	GeLU dropout
	& \multicolumn{7}{c}{0.0} \\
	Attention dropout
	& \multicolumn{7}{c}{0.1} \\
	Max sequence length
	& 512 & 128 & 128 & 512 & 512 & 512 & 512 \\
	Batch size
	& 32 & 32 & 64 & 128 & 128 & 128 & 128 \\
	Number of epochs
	& 5 & 3 & 3 & 3 & 3 & 3 & 3 \\
	Learning rate decay & \multicolumn{7}{c}{Linear} \\
	Weight decay        & \multicolumn{7}{c}{0.01}   \\
	Warmup proportion   & \multicolumn{7}{c}{0.1}   \\
	Adam epsilon        & \multicolumn{7}{c}{1e-6}   \\
	\bottomrule
\end{tabular}
\caption{Hyper-parameters for finetuning on the GLUE benchmark and 7 text classification datasets.}
\label{tab:hp-finetune-glue-textcls}
\end{table}
For all the finetuning experiments, we essentially inherit the hyper-parameters used by XLNet~\cite{yang2019xlnet}.
All the performance numbers reported are obtained on TPUs with TensorFlow 2.2.

\subsubsection{GLUE \& Text Classification}
For GLUE and text classification datasets, we first fix the values of most hyper-parameters shown in Table~\ref{tab:hp-finetune-glue-textcls}.
Then, we only search the learning rates from the set [1e-5, 2e-5, 3e-5], and choose the best one according to the validation set.

Following previous work~\cite{yang2019xlnet,liu2019roberta,clark2020electra}, all GLUE performances correspond to the median result of 5 runs from different random seeds in the base setting and 15 runs in the large setting, respectively.

For the text classification, the base-scale results are the median performance among 5 runs with different random seeds.
However, for the large-scale experiments, to be compatible with previous work~\cite{xie2019unsupervised,yang2019xlnet}, the results are the best performance among 5 random runs.

\subsubsection{Reading Comprehension}
Again, following XLNet~\cite{yang2019xlnet}, the hyper-parameters used for finetuning on the RACE and SQuAD datasets are summarized in Table~\ref{tab:hp-finetune}.
``Layer-wise decay'' means exponentially decaying the learning rates of individual layers in a top-down manner.
For example, suppose the $24$-th layer uses a learning rate $l$, and the Layer-wise decay rate is $\alpha$, then the learning rate of layer $m$ is $l \alpha^{24 - m}$.
In addition, for the two versions of SQuAD, we simply reuse the model trained on SQuAD v2.0 when evaluated on SQuAD v1.1.
\begin{table}[!h]
\centering
\begin{tabular}{lcc}
	\toprule
	\bf Hparam & \bf RACE & \bf SQuAD \\
	\midrule
	Dropout                 & \multicolumn{2}{c}{0.1}    \\
	Attention dropout       & \multicolumn{2}{c}{0.1}    \\
	Max sequence length     & 512          & 512 \\
	Training epochs/steps   & 5 epochs     & 8000 steps \\
	Warmup proportion/steps & 0.1          & 1000 steps \\
	Batch size              & [16, 32]     & 48 \\
	Learning rate           & [1e-5, 2e-5] & 3e-5 \\
	Learning rate decay     & \multicolumn{2}{c}{linear} \\
	Weight decay            & \multicolumn{2}{c}{0.01}   \\
	Adam epsilon            & \multicolumn{2}{c}{1e-6}   \\
	Layer-wise lr decay     & 1.0 & 0.75 \\
	\bottomrule
\end{tabular}
\caption{Hyper-parameters for RACE and SQuAD.}
\label{tab:hp-finetune}
\end{table}

\section{Additional Experimental Results}
\label{sec:appendix-extra-exp}

\subsection{Text Classification at Large Scale}
\label{sec:appendix-textcls}
\begin{table}[!ht]
	\centering
	\begin{tabular}{lccccccc}
		\toprule
		\bf Model & \bf IMDB & \bf AG & \bf DBpedia & \bf Yelp-2 & \bf Yelp-5 & \bf Amazon-2 & \bf Amazon-5 \\
		\cmidrule(r){1-1} \cmidrule(l){2-8}
		BERT-Large
		& 4.51 & - & 0.64 & 1.89 & 29.32 & 2.63 & 34.17 \\
		ROBERTA-Large
		& 3.50 & - & -    & -    & -     & -    & -     \\
		XLNet-Large
		& \bf 3.20 & \bf 4.45 & 0.64 & 1.37 & \bf 27.05 & 2.11 & 31.67 \\
		B10-10-10H1024
		& 3.36 & 4.66 & \bf 0.60 & \bf 1.33 & 27.14 & \bf 2.10 & \bf 31.64 \\
		B8-8-8H1024
		& 3.42 & 4.96 & 0.63 & 1.39 & 27.20 & 2.14 & 31.74 \\
		\cmidrule(r){1-1} \cmidrule(l){2-8}
		MPNet
		& 4.40 & -  & -     & -   & -   & -     & -  \\
		B6-6-6H768
		& \bf 3.72 & \bf 5.00 & \bf 0.64 & \bf 1.50 & \bf 27.73 \bf & \bf 2.27 & \bf 32.11 \\
		B6-3x2-3x2H768
		& 3.82 & 5.12 & 0.64 & 1.58 & 27.96 & 2.32 & 32.23 \\
		B4-4-4H768
		& 4.12 & 5.09 & 0.67 & 1.70 & 28.40 & 2.35 & 32.46 \\  
		\bottomrule 
	\end{tabular}
	\caption{Text classification performance comparison under the large-scale pretraining.}
	\label{tab:large-scale-textcls}
\end{table}
Table \ref{tab:large-scale-textcls} includes the performance comparison on 7 text classification tasks under the large-scale training setting.
Similar to the GLUE benchmark results, compared with the previous result based on Transformer, with fewer FLOPs, the proposed F-TFM achieves comparable results.

\subsection{Training Cost Comparison}
\label{sec:appendix-training-cost}
In this section, we test the pretraining and finetuning speed of the F-TFM in comparison to the standard Transformer on the TPU and GPU platform.
For the pretraining speed evaluation, we test F-TFM on TPU v3-16 (16 cores x 16Gb) with TensorFlow. 
For the finetuning speed evaluation, we test F-TFM on TPU v2-8 (8 cores x 8Gb) with TensorFlow and on Nvidia-V100 (16Gb) GPU with the PyTorch.
The TensorFlow version is 2.2.0, and the PyTorch version is 1.5.0.
For the GPU experiments, we use an 8-GPU node on the Google Cloud Platform.
All running speeds are reported with the FP16 optimizer.
In the PyTorch implementation, we use ``O2'' options of AMP manager in the apex\footnote{\url{https://github.com/NVIDIA/apex}} package to handle the FP16 optimization.
For finetuning, we consider three different sequence lengths, namely 128, 256 and 512.
For pretraining, we only consider the sequence length 512.
In each case, we choose the maximum possible batch size allowed by the memory size of the device(s). 
We measure the actual model \textit{running time} by performing 1000 steps gradient descent with random input sequences with the fixed length. 

\begin{table}[!ht]
	\small
	\centering
	\begin{tabular}{@{}l|cc|c|cc|c|c|c|c@{}}
		\toprule
		Sequence length & \multicolumn{3}{c|}{128} & \multicolumn{3}{c|}{256} & \multicolumn{2}{c|}{512} & \\
		\midrule
		\multirow{2}{*}{Metrics}
		& \multicolumn{2}{c|}{Run time} & \multirow{2}{*}{Mem} 
		& \multicolumn{2}{c|}{Run time} & \multirow{2}{*}{Mem} 
		& \multicolumn{1}{c|}{Run time} & \multirow{2}{*}{Mem} 
		& \multirow{2}{*}{GLUE} \\
		& 1 GPU & 8 GPUs &  & 1 GPU & 8 GPUs & & 8 GPUs & &  \\
		\midrule\midrule
		Batch size / GPU & \multicolumn{3}{c|}{64} & \multicolumn{3}{c|}{32} & \multicolumn{2}{c|}{16} & \\
		\cmidrule(r){1-1} \cmidrule(lr){2-4}\cmidrule(lr){5-7}\cmidrule(lr){8-9} \cmidrule(l){10-10}
		L12H768
		& 1.00x & 1.00x & 9.2G & 1.00x & 1.00x & 11.0G & 1.00x & 14.3G & 84.40 \\
		B6-6-6
		& 0.97x & 0.99x & 9.1G & 0.95x & 0.97x & 10.3G & 0.94x & 12.5G & 85.37 \\
		B6-3x2-3x2
		& 0.93x & 0.93x & 8.4G & 0.91x & 0.92x & 9.5G & 0.90x & 11.8G & 84.78 \\
		B4-4-4
		& 0.67x & 0.67x & 6.6G & 0.65x & 0.66x & 7.5G & 0.64x & 9.0G & 83.99 \\
		\midrule\midrule
		Batch size / GPU & \multicolumn{3}{c|}{32} & \multicolumn{3}{c|}{12} & \multicolumn{2}{c|}{4} & \\
		\cmidrule(r){1-1} \cmidrule(lr){2-4}\cmidrule(lr){5-7}\cmidrule(lr){8-9} \cmidrule(l){10-10}
		L24H1024
		& 1.00x & 1.00x & 14.8G & 1.00x & 1.00x & 14.4G & 1.00x & 13.9G & 86.62 \\
		B10-10-10
		& 0.87x & 0.92x & 14.0G & 0.90x & 0.93x & 13.0G & 0.96x & 12.7G & 87.03 \\
		B8-8-8
		& 0.70x & 0.73x & 11.6G & 0.73x & 0.75x & 10.8G & 0.78x & 10.5G & 86.70 \\
		\bottomrule
	\end{tabular}
	\caption{
		Running time and memory consumption comparison between F-TFMs and the standard Transformer on the GPU. 
		In each model group, the standard Transformer (first model) is used as the benchmark for the rest of F-TFM models.
		Note that, given the same batch size per GPU, the memory consumption is roughly the same for 1 GPU and 8 GPUs.
	}
	\label{tab:gpu-time}
\end{table}

\begin{table}[!ht]
\centering
\begin{tabular}{l|c|c|c|c}
	\toprule
	Sequence length & \hspace{0.2in} 128 \hspace{0.2in} & \hspace{0.2in} 256 \hspace{0.2in} & \hspace{0.2in} 512 \hspace{0.2in} & \\
	\midrule
	Metrics
	& \multicolumn{3}{c|}{Run time on 8 TPU cores (TPUv2-8)} & GLUE \\
	\midrule\midrule
	Batch size / TPU core & 64 & 32 & 16 & \\
    \cmidrule(r){1-1}\cmidrule(lr){2-2}\cmidrule(lr){3-3}\cmidrule(lr){4-4}\cmidrule(l){5-5}
	L12H768
	& 1.00x & 1.00x & 1.00x & 84.40 \\
	B6-6-6
	& 0.99x & 0.88x & 0.81x & 85.37 \\
	B6-3x2-3x2
	& 0.97x & 0.87x & 0.77x & 84.78 \\
	B4-4-4
	& 0.69x & 0.62x & 0.55x & 83.99 \\
	\midrule\midrule
	Batch size / TPU core & 16 & 8 & 4 & \\
	\cmidrule(r){1-1}\cmidrule(lr){2-2}\cmidrule(lr){3-3}\cmidrule(lr){4-4}\cmidrule(l){5-5}
	L24H1024
	& 1.00x & 1.00x & 1.00x & 86.62 \\
	B10-10-10
	& 0.89x & 0.81x & 0.73x & 87.03 \\
	B8-8-8
	& 0.66x & 0.60x & 0.56x & 86.70 \\
	\bottomrule
	\end{tabular}
	\caption{
		Running time between F-TFMs and the standard Transformer on the TPU v2-8. 
		In each model group, the standard Transformer (first model) is used as the benchmark for the rest of F-TFM models.
	}
	\label{tab:tpu-time}
\end{table}

Firstly, we compare the model speed in the finetuning stage. 
Note that the decoder is not used in this setting.
Table~\ref{tab:gpu-time} and~\ref{tab:tpu-time} summarize the finetuning running time comparison on GPUs and TPUs, respectively. 
\begin{itemize}[leftmargin=*,itemsep=0em]
\item In the base model (L12H768) group, we observe that the speed of B6-6-6H768 is similar or faster than the base Transformer model, despite the fact that B6-6-6 is deeper, has more parameters. 
Moreover, B6-6-6H768 achieves better results compared with the base Transformer model.
The similar conclusion applies to the B6-3x2-3x2 model, which has the same amount of parameters as the base model.
The B4-4-4 model, which has the same depth and model parameters as the base model, is able to provide 30\%-50\% speedup without losing too much performance. 
\item In the large model (L24H1024) group, the conclusion is similar. The speed of the larger model B10-10-10 is almost the same as the large model, and the speed of B8-8-8 is significantly faster than the large model. In addition, when sequence length equals 512, the acceleration of F-TFM on the TPU  is more obvious than the GPU.
\item In the both groups, all the tested F-TFM variants have smaller memory footprint compared with the standard TFM models, showing the memory efficiency of F-TFM. 
\end{itemize}

Next, we compare the model speed during pretraining under the MLM objective in table \ref{tab:tpu-pre-time}, which has an additional cost due to the decoder.
The results show that the proposed method can still substantially improve the pretraining speed compared to the standard Transformer, though the speed gain is slightly smaller than the finetuning stage.
In summary, this study demonstrates that the proposed method is more efficient in both the finetuning and pretraining stages in modern parallel computing platforms. 
\begin{table}[!ht]
\centering
\begin{tabular}{lcc}
	\toprule
	Sequence Length
	& \multicolumn{2}{c}{512} \\
	\cmidrule(r){1-1} \cmidrule(l){2-3}
	& Running Time & FLOPs \\
	\cmidrule(r){1-1} \cmidrule(l){2-3}
	\#TPU cores / Total bsz 
	& \multicolumn{2}{c}{16 / 512} \\
	\cmidrule(r){1-1} \cmidrule(l){2-3}
	L12H768          & 1.00x & 1.00x \\
	B6-6-6H768D2     & 0.99x & 1.04x \\
	B6-3x2-3x2H768D2 & 0.97x & 1.04x \\
	B4-4-4H768D2     & 0.79x & 0.75x \\
	\midrule
	\#TPU cores / Total bsz
	& \multicolumn{2}{c}{16 / 128} \\
	\cmidrule(r){1-1} \cmidrule(l){2-3}
	L24H1024         & 1.00x & 1.00x \\
	B10-10-10H1024D2 & 0.83x & 0.81x \\
	B8-8-8H1024D2    & 0.71x & 0.66x \\
	\bottomrule
\end{tabular}
\caption{TPU pretraining speed comparison. The suffix ``D2'' means that the F-TFM model has 2 decoder layers.} 
\label{tab:tpu-pre-time}
\end{table}

\end{document}